\documentclass[conference]{IEEEtran}
\usepackage{cite}

\ifCLASSINFOpdf
  \usepackage[pdftex]{graphicx}
\else
\fi
\usepackage{amsmath}
\usepackage{array}

  \usepackage[caption=false,font=footnotesize]{subfig}
\usepackage{url}

\usepackage{pgfplots}
\usepackage{amssymb}
\usepackage{booktabs}
\usepackage{balance}

\newcommand{\Naz}{{\bf Nazr-CNN}}

\begin{document}
%
\title{Nazr-CNN: Fine-Grained Classification of UAV Imagery for Damage Assessment}



%
\author{\IEEEauthorblockN{Nazia Attari\IEEEauthorrefmark{1},
Ferda Ofli,
Mohammad Awad, 
Ji Lucas and
Sanjay Chawla}
\IEEEauthorblockA{Qatar Computing Research Institute\\
Hamad bin Khalifa University,
Doha, Qatar
}
\IEEEauthorblockA{
Email: \IEEEauthorrefmark{1}nazia.ec@gmail.com,  
\{fofli, mhasan, jlucas, schawla\}@hbku.edu.qa}
}


\maketitle

\begin{abstract}
We propose \Naz\footnote{Nazr means ``sight'' in Arabic.}, a deep learning pipeline for object detection and fine-grained classification in images acquired from Unmanned Aerial Vehicles (UAVs) for damage assessment and monitoring. \Naz\ consists of two components. The function of the first component is to localize objects (e.g. houses or infrastructure) in an image by carrying out a pixel-level classification. In the second component, a hidden layer of a Convolutional Neural Network (CNN) is used to encode Fisher Vectors (FV) of the segments generated from the first component in order to help discriminate between different levels of damage. 

To showcase our approach we use data from UAVs that were deployed to assess the level of damage in the aftermath of a devastating cyclone that hit the island of Vanuatu in 2015. The collected images were labeled by a crowdsourcing effort and the labeling categories consisted of fine-grained levels of damage to built structures. 
Since our data set is relatively small, a pre-trained network for pixel-level classification and FV encoding was used. \Naz\ attains promising results both for object detection and damage assessment suggesting that the integrated pipeline is robust in the face of small data sets and labeling errors by annotators. While the focus of \Naz\ is on assessment of UAV images in a post-disaster scenario, our solution is general and can be applied in many diverse settings. We show one such case of transfer learning to assess the level of damage in aerial images collected after a typhoon in Philippines.
\end{abstract}


%
\IEEEpeerreviewmaketitle

\section{Introduction}

Unmanned Aerial Vehicles (UAVs) are now being increasingly used for 
structural damage assessment during routine monitoring and in
the aftermath of natural disasters. For example, the use of UAVs for monitoring
electrical power lines is gaining prominence, and so is their importance
for damage inspection post a natural disaster event~\cite{galarette}.
In fact, both the United States Federal Emergency Management Agency (FEMA) and the European Commission's Joint Research Center (JRC) have noted that aerial imagery will play an important role in disaster response and present a big data challenge~\cite{Ofli:BDJ16}.

As a specific example, the World Bank cooperated with the Humanitarian UAV Network (UAViators)\footnote{http://uaviators.org/} in the wake of Cyclone Pam, a category five cyclone that caused extensive damage in Vanuatu in March 2015, to carry out a post-disaster assessment of Vanuatu. The workflow followed was analogous to that of AIDR\footnote{http://aidr.qcri.org/} (Artificial Intelligence for Disaster Response~\cite{Imran}), a system developed by QCRI to harness information from real-time tweets collected from an area struck by a natural disaster to help coordinate humanitarian relief activities. The acquired UAV images were annotated by a group of volunteers using MicroMappers\footnote{http://www.micromappers.org/}, which is a crowdsourcing platform also built by QCRI in partnership with United Nations and the Standby Task Force, specifically for crisis management. Each annotator was asked to demarcate houses using a polygonal region and then associate a label with each region indicating the severity of damage (i.e., mild, medium and severe).


\begin{figure}
    \centering
    \includegraphics[width=.95\columnwidth]{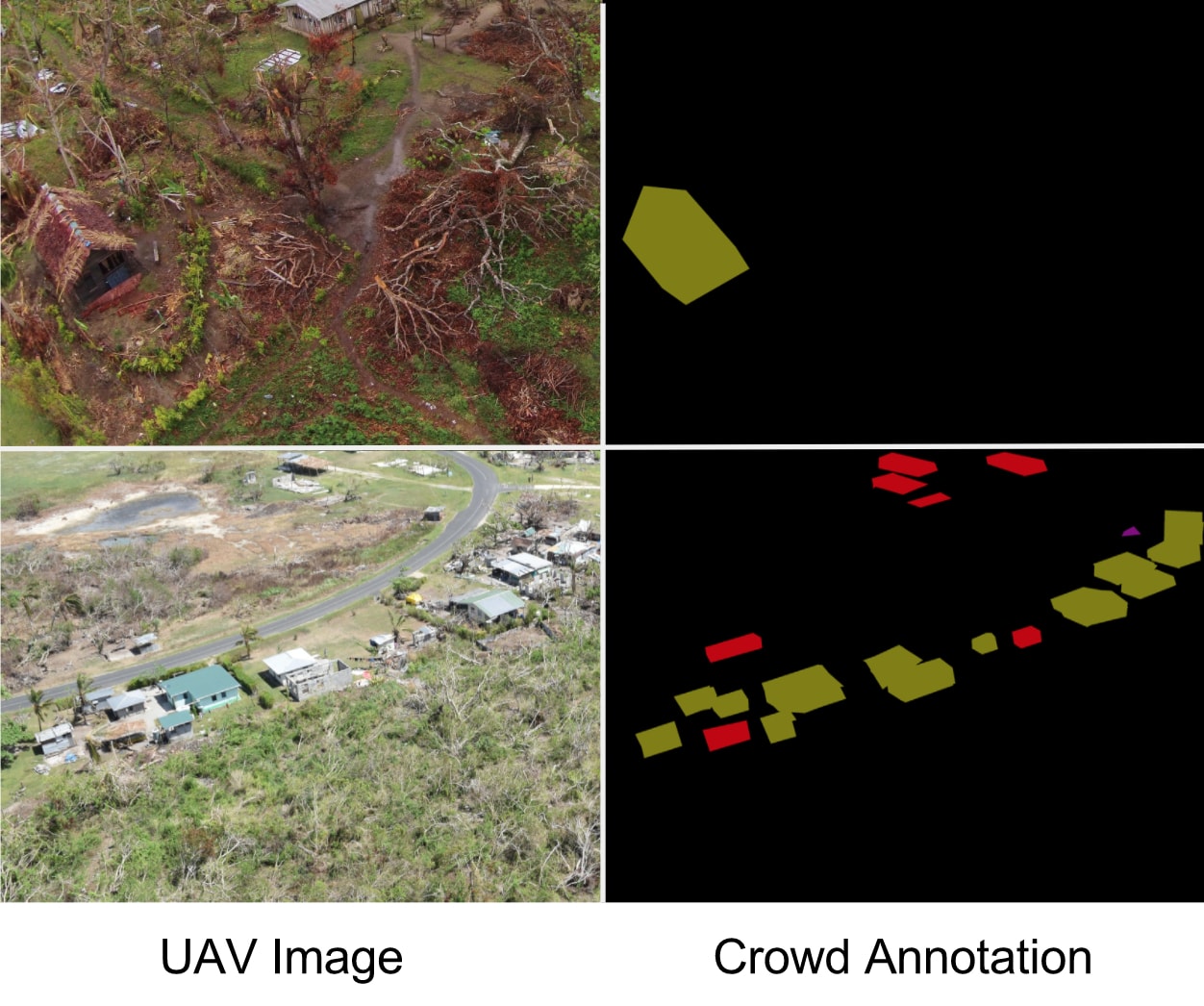}\\
    \includegraphics[width=.8\columnwidth]{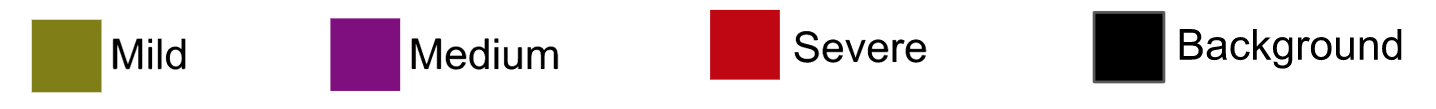}
    \caption{ Disaster images at different angle and elevation. Left: images-from-UAV. Right: Crowd annotations. Notice the difficulty of correctly annotating the images and thus generating accurate ground truth.}
    \label{fig:hetero1}
\end{figure}

Analyzing large volumes of high-resolution aerial images generated after a major disaster remains  a challenging task in contrast to the ease of acquiring them due to low operational costs. A popular approach is to use a hybrid strategy where initially a crowdsourcing effort is carried out to create a labeled training set which is used to infer a  machine learning (ML) model~\cite{Imran}. The ML model then automatically classifies incoming images. However, until now, the image classification task (unlike text classification) suffered from low accuracy and lack of robustness in an uncontrolled environment.

Before we describe our solution, we enumerate some of the challenges that must be overcome for object detection and fine-grained classification in images acquired from UAVs:
\begin{enumerate}
\item
A single UAV image usually contains a high number of objects at different scales belonging to multiple categories. Also, the background in the image is often highly heterogeneous, e.g., ocean, sky, forest, grassland, etc. In Figure~\ref{fig:hetero1}, we show a couple of example UAV images from our data set which confirm not only the difficulty of object detection task but also challenges in annotation. In the top image, the shack is almost indistinguishable from the land both in terms of texture and color.
\item
The amount of labeled UAV imagery data is limited and the labels are often noisy since the damage assessment task is profoundly subjective. Furthermore, labeling small objects whose damage type is difficult to gauge makes the annotation task a lot more tedious. In our particular case, there is a significant conflict between built structures whose damage levels are rated as \textit{medium} or \textit{severe} (see Figure~\ref{fig:labdis} for example images).
\item
A key success of deep learning is that ``feature engineering'' is part of the learning process. However, we have observed, at least in our data set, distinguishing images based on texture is an important aspect of the problem and straightforward application of CNN is unlikely to result in fine-grained object classification.
\end{enumerate}

\subsection{Solution in a Nutshell}
Our solution (\Naz) combines two deep learning pipelines. The first pipeline carries out a pixel-level classification (often known as semantic segmentation) to localize objects (segments) in images. The localized objects are then passed through a pre-trained Convolutional Neural Network (CNN) and a Fisher Vector (FV) encoding is extracted from the single (last) convolutional layer to generate a new representation of the objects. The FVs are then trained using a standard SVM classifier. This results in a highly accurate detection and fine-grained discrimination of houses based on their damage levels.

The rest of the paper is structured as follows. In Section~\ref{sec:probdef} we precisely define and state the problem of fine-grained object classification. In Section~\ref{sec:deeplearning}, we introduce the building blocks of \Naz\ with a particular focus on the use of FV encoding. In Section~\ref{sec:exp_res}, we present the experiments and results. We survey the related work in Section~\ref{sec:rel_work}, and conclude the paper with a summary and future directions in Section~\ref{sec:conclusion}.

\section{Problem Statement}
\label{sec:probdef}
We now define the problem of object detection and classification in
UAV images for damage assessment. \\[1ex]
\noindent
{\bf Given:} A set of images obtained from  UAVs over a disaster assessment area and labeled
using a crowdsourcing platform. Built structures (e.g. houses) are labeled as {\it little-to-no damage/mild (M), partially damaged/medium (Md) and fully destroyed/severe (S).} \\[1ex]
{\bf Design:} A machine learning classifier which takes as input an unlabeled image and
classifies regions in the image as background (B) or containing structures which
can be classified as M, Md or S. \\[1ex]
{\bf Constraints:} 
(i)  Images often contain large number of objects with different damage types; (ii) the size of the data set is relatively small and (iii) there is disagreement among the annotators on the class labels.

\section{Deep Learning Framework}
\label{sec:deeplearning}
Since our labeled data set is relatively small (1,085 images), we have built our deep learning pipeline (\Naz) using existing pre-trained networks as building blocks. Our proposed pipeline consists of integrating pixel-level classification (often known as semantic image segmentation) and texture discrimination. For semantic segmentation we have used DeepLab~\cite{deeplab:iclr15}, a CNN network followed by a fully-connected Conditional Random Field (CRF) for smoothing. For texture discrimination we have used FV-CNN~\cite{cimpoi2015deep}, which consists of extracting
Fisher Vectors from a  hidden layer of a pre-trained CNN.
Intuitively, the aim of semantic segmentation is to localize objects (i.e., houses) in an image and the aim of texture discrimination is to distinguish between different types of damage and background. For a comprehensive background on CNNs, we refer the reader to the upcoming book by the pioneers of the field~\cite{goodfellow}.

\subsection{Pre-Trained Networks}
In practice, CNNs are rarely fully trained afresh for new data sets because it is relatively rare to have access to large data sets--which is indeed the case we have.  A popular choice of a pre-trained network is the VGG-16 network~\cite{simonyan2014very} that is trained on the ImageNet data set which contains over 1.2 million images and 1,000 categories (labels). The VGG-16 network consists of 16 layers and over 140 million weight parameters. There are two ways that a pre-trained network is used on new data sets. The first approach is to just pass each data point from the new data set and use the layers as feature extractors where each data point can be mapped into a new representation. Lower layers of VGG-16 can be considered as low-level feature extractors which should be applicable across domains. Higher layers tend to be more domain-specific. The second approach is to use the existing weights of the pre-trained network as an initialization for the new data set. This approach can often prevent overfitting but is computationally expensive.

\subsection{Semantic Segmentation}
We have used the DeepLab~\cite{deeplab:iclr15} to carry out pixel-level classification of images (i.e., semantic segmentation). This promotes the localization of objects in images. DeepLab combines the VGG-16 network with a fully-connected Conditional Random Field (CRF) model on the output of the final layer of CNN.  The CRF overcomes the poor localization property of CNNs and results in better segmentation. The CRF optimizes the following energy function:
\begin{equation}
E(\mathbf{x}) = \sum_{i}g_{i}(x_{i}) + \sum_{ij}h_{ij}(x_{i},x_{j})
\end{equation}
where $\mathbf{x}$ is the pixel-level label assignment and $g_{i} = -\log P(x_{i})$ is the label assignment probability at pixel $i$ computed by CNN.  
The pairwise potential is given by
\[
h_{ij}(x_{i},x_{j}) = \mu(x_{i},x_{j})\sum_{m=1}^{K}w_{m}\dot k^{m}(\mathbf{f_{i}},\mathbf{f_{j}})
\]
Here $\mu$ is the binary Potts model function, $k^{m}$ is a Gaussian kernel for label $k$ and is dependent on the features $\mathbf{f}$ extracted at the pixel level. The CRF is fully connected, i.e., there is one pair-wise term for each pair of pixels irrespective of whether they are neighbors or not.  The reason that DeepLab uses fully connected CRF is that the objective is to extract local structure (shape) from the pixels and not just carry out a local smoothing (which might smooth out the local shapes). Pixel-level classification is an integral part of \Naz. Besides identifying the houses in UAV images and their shapes, pixel-level classification, as we will see in Section~\ref{sec:exp_res}, serves as an automatic data cleaning step: it is robust against annotation errors of both kinds, i.e., it can identify segments which were missed by the annotator as well as fix labeling errors.

\subsection{Fisher Vector Encoding}
We use FV-CNN~\cite{cimpoi2015deep} to extract features which help in distinguishing between different levels of damage. In particular, we explain why the use of FVs as a representation of the input can help distinguish between different levels of damage. Since our data set is of modest size (1,085 labeled images), as is the usual practice, we use FV-CNN based on VGG-M architecture pre-trained on the ImageNet ILSVRC 2012 data set.

Fisher Vectors (FV) are a generalization of the popular Bag-of-Visual words (BoV) representation of images and are known to result in substantial increase in accuracy for image classification tasks~\cite{sanchez2013image}. As we will show that FVs extracted from high level CNN features are particularly useful for distinguishing different types of building damages. 

We assume that an appropriate layer of CNN will generate a set of features $X = \{x_{i}, i= 1,\ldots, N\}$ where each $x_{i} \in \mathbb{R}^{D}$. Further we assume that the set $X$ is generated from a Gaussian Mixture Model (GMM). Thus for each $x \in X$,
\begin{equation}
 P(x|\lambda) = \sum_{i=1}^{K}w_{i}N(x,\mu_{i},\Sigma_{i}) 
\end{equation}
and 
\begin{equation}
\forall i \;: w_{i} \geq 0, \hspace{1cm} \sum_{i=1}^{K}w_{i} = 1
\end{equation}
Here $N(x,\mu,\Sigma)$ is a multi-dimensional Normal (Gaussian) distribution. In order to avoid enforcing the constraints on the weights ($w$), a re-parameterization is carried out such that 
\begin{equation}
w_{i}  = \frac{exp(\alpha_{i})}{\sum_{j=1}^{K}exp(\alpha_{j})}
\end{equation}
Now the FV encoding of an $x \in X$ are the gradients of $\log P(x|\lambda)$ with respect to $\lambda =\{\alpha_{j},\mu_j,\Sigma_{j}, j=1,\ldots K\}$. The covariance matrix $\Sigma_{j}$ is assumed to be a diagonal matrix. Thus
\begin{align}
\nabla_{\alpha_{j}}\log P(x) & =  \gamma(j) - w_j  \label{weights}\\
\nabla_{\mu_{j}}\log P(x) & =  \gamma(j)\left(\frac{x - \mu_{j}}{\sigma_{j}^{2}}\right) \label{meansss} \\
\nabla_{\sigma_{j}}\log P(x) & =  \gamma(j)\left[\frac{(x - \mu_{j})}{\sigma_{j}^{3}} - \frac{1}{\sigma_{j}}\right] \label{variances}
\end{align}
where the responsibility $\gamma(j)$ (the posterior probability) that $x$ belongs to $N(u_{j},\Sigma_{j})$ is given by
\begin{equation}
\gamma(j) = \frac{w_jN(x,u_{j},\Sigma_{j})}{\sum_{i=1}^{K}w_{i}N(x,u_{i},\Sigma_{i})}
\end{equation}

Often further normalization is carried out in Equations~\ref{weights},~\ref{meansss}, and~\ref{variances} to  arrive at the  precise  Fisher Vectors. Further details are provided in ~\cite{sanchez2013image}.

\begin{figure}
\centering
\includegraphics[scale=0.5]{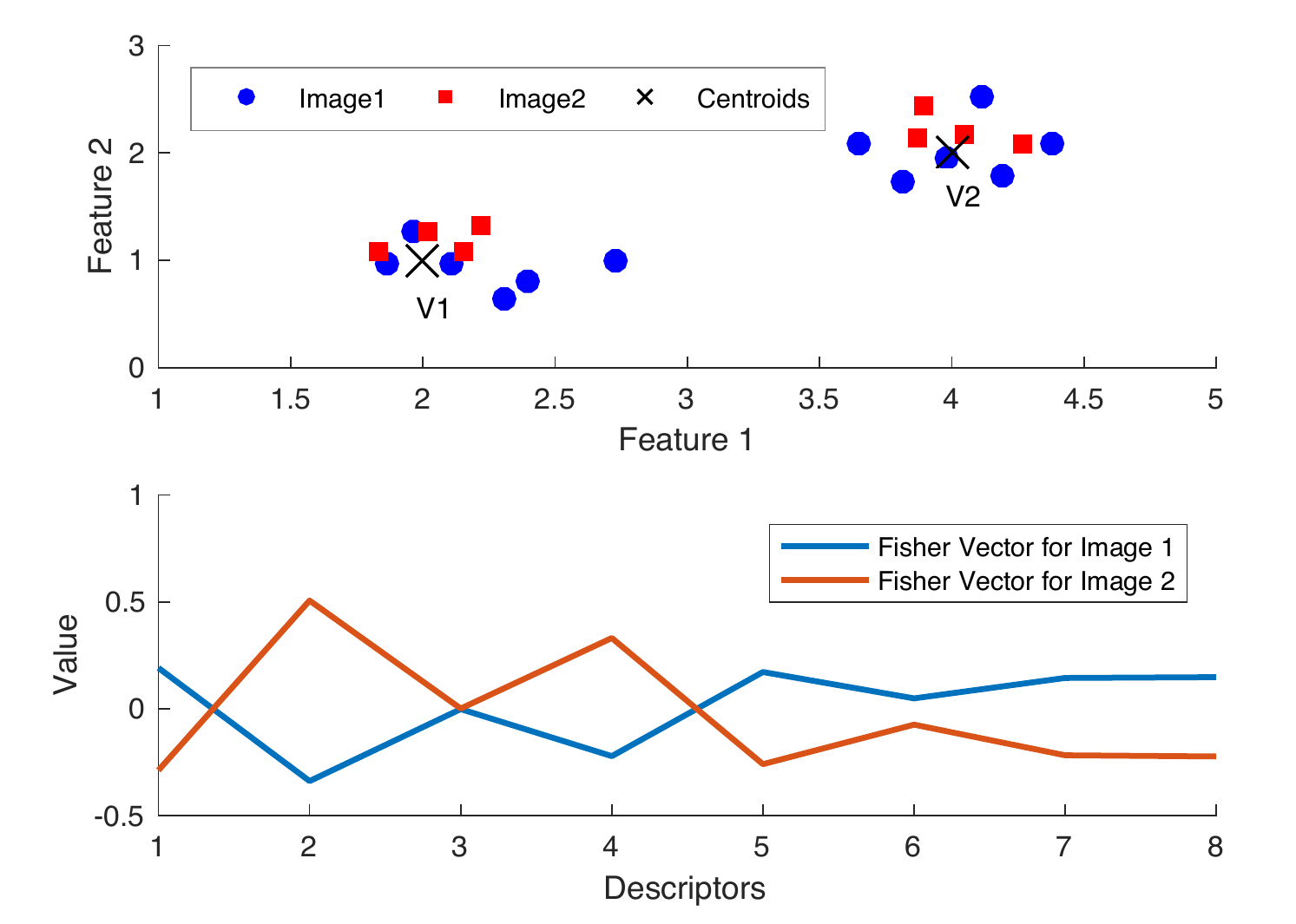}
\caption{Bag of Visual Words (BoV) vs. Fisher Vector (FV) representation.}
\label{bov_fish}
\end{figure}

\smallskip
\noindent{\bf Example:} We now give a simple example to show why FV encoding results in more discriminative  features compared to the BoV model. Consider the example show in Figure~\ref{bov_fish}.  Assume we are given two-dimensional descriptors of two images (red and blue in the figure). The BoV model is to cluster the descriptors using the k-means algorithm and use the centroid as a representation of a visual word. In the example there are two visual words. Then each image is represented as a histogram consisting of counts associated with the visual words. For example the histogram of the blue image is $(6,6)$ as six descriptors of the blue image are associated with the first visual word and six with the second visual word.

In contrast, the FV of the image has a dimensionality equal to the number of parameters of the GMM. For example the bottom plot in Figure~\ref{bov_fish} shows a parallel plot of eight of the ten features of each image. The FV is more sensitive to local variations (as the value at each descriptor represents the deviation from the GMM model). This is particularly suitable for texture discrimination where variation within a region and not the shape of the region is an important parameter.

\subsection{Proposed Pipeline}
\begin{figure}
\centering
\includegraphics[width=\columnwidth]{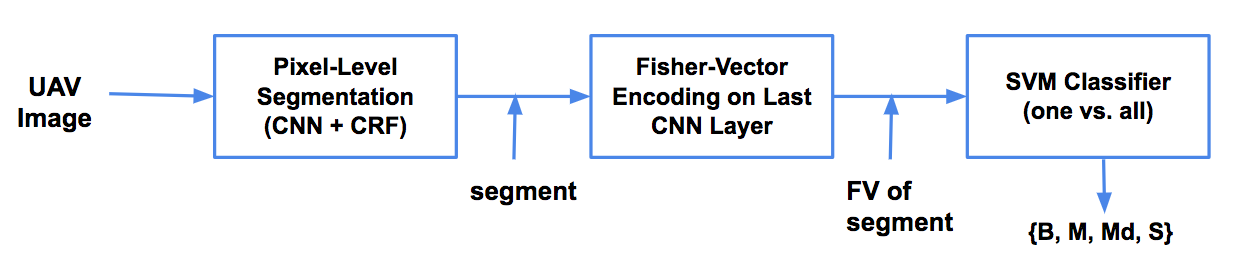}
\caption{\Naz\ combines pixel-level classification with FV-CNN. The Fisher Vectors are then trained using multi-class SVM.}.
\label{naz}
\vspace{-5mm}
\end{figure}

We now present our fine-grained image classification pipeline (\Naz) for damage assessment in UAV images. Figure~\ref{naz} shows the work flow of the proposed pipeline.

\noindent{\bf Training:}
\begin{enumerate}
\item For pixel-level classification, the DeepLab system requires the image along with the masks created by the annotators.
\item The DeepLab system generates segments. Each segment is assigned with a label based on the ground truth mask with the maximum overlap.
\item The segment output from DeepLab is then fed into FV-CNN. For each segment, Fisher Vectors from an intermediate hidden layer of FV-CNN will be generated. Along with the Fisher Vectors, the label of the segment forms an element of the training set of the multi-class SVM. An SVM model is then induced from the segment and its label.
\end{enumerate}

\noindent{\bf Testing:}
\begin{enumerate}
\item An image (without annotation) is passed through DeepLab to generate segments.
\item Each segment is fed into FV-CNN, which generates Fisher Vectors for the segments.
\item The Fisher Vector of each segment is classified by the SVM model into one of the damage categories, i.e., B, M, Md, S.
\end{enumerate}

\section{Experiments and Results}
\label{sec:exp_res}
In this section, we report on the extensive set of experiments that we have carried out to assess the performance of \Naz. We begin by describing in Section~\ref{ssec:data_description} the data acquisition process used in our experiments. Then, we report the following four results: (i) the baseline accuracy of FV-CNN on pre-labeled segments to discriminate between severity of damage (Section~\ref{ssec:fvcnn_baseline}), (ii) the baseline accuracy of DeepLab on ground truth annotations to localize and classify damaged houses (Section~\ref{ssec:deeplab_baseline}), and (iii) the accuracy of \Naz\ which is the combined pipeline of pixel-level semantic segmentation (with cross-validation) and Fisher Vector encoding using FV-CNN along with precision-recall computation (Section~\ref{ssec:proposed_res}), and finally, (iv) the result of transfer learning on a novel dataset (Section~\ref{ssec:transfer_res}).

\subsection{Data Description}
\label{ssec:data_description}
\begin{figure}
\centering
\includegraphics[width=0.85\columnwidth]{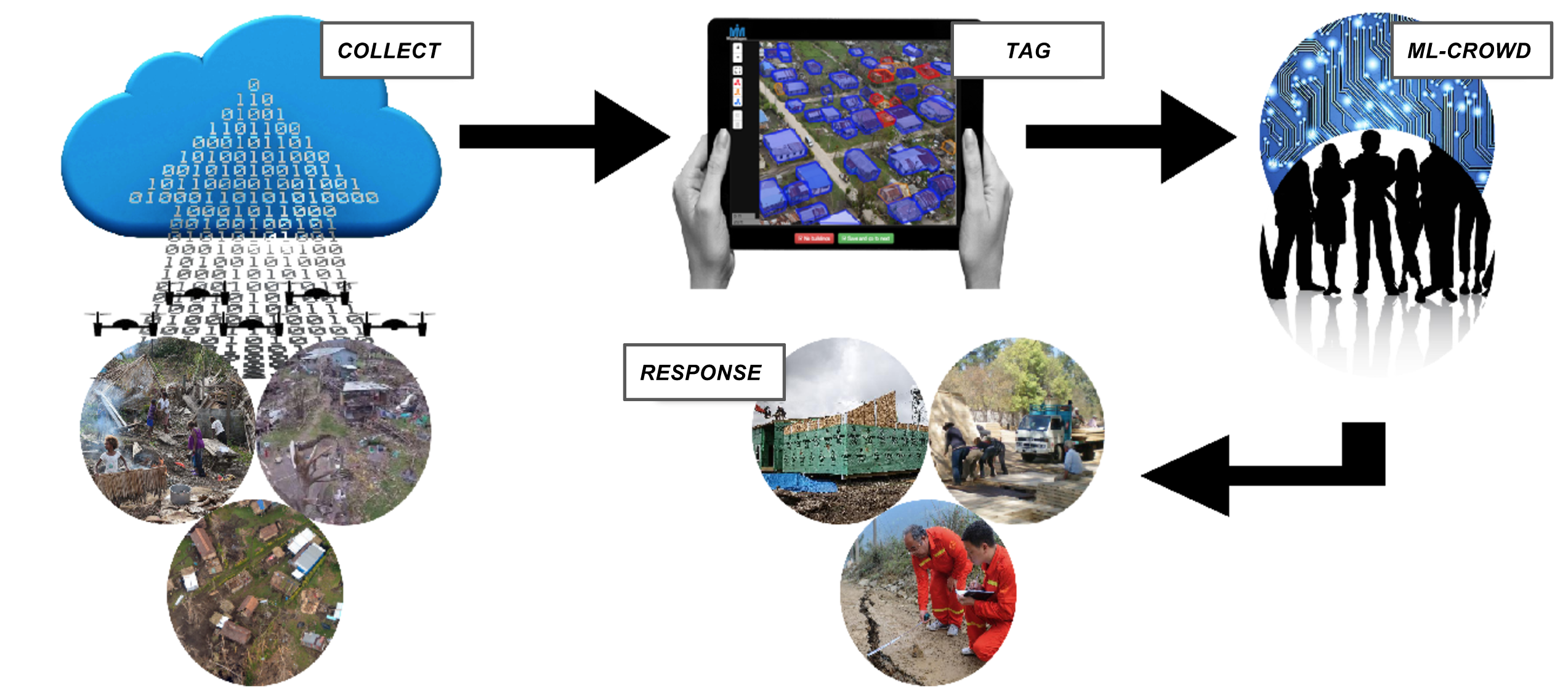}
\caption{UAV image acquisition and annotation workflow implemented in MicroMappers for this study, following the approach used for text classification by AIDR\protect\footnotemark. However, annotation of disaster images is substantially more difficult and until now, there did not exist an accurate and robust object detection and classification model.}
\label{hum_workflow}
\vspace{-5mm}
\end{figure}

The UAV image data and corresponding damage annotations were acquired as part of an initiative by the World Bank
in collaboration with the Humanitarian UAV Network (UAViators) during Cyclone Pam in Vanuatu in 2015. The workflow followed was analogous to that of AIDR (Artificial Intelligence for Disaster Response~\cite{Imran}) for text data, and is shown in Figure~\ref{hum_workflow}: images were acquired through UAVs and a group of digital volunteers using MicroMappers annotated built structures found in the images as described below.
The resulting image dataset contains 3,096 images but approximately 65\% of them do not contain any built structures or ground truth annotations. Hence, we use only a set of 1,085 images where each image contains one or more regions with different levels of damage: Mild (M), Medium (Md), Severe (S); and everything else is considered as Background (B).
In addition to this data set, we have 60 images from a typhoon disaster that affected areas in Philippines to test the generalization potency of \Naz.

\footnotetext{Because of the cyclone the Internet was down and thus AIDR could actually not be activated to process tweets.}

\subsubsection{UAV Image Annotation} Images have been labeled using an online annotation platform and polygons are drawn around each built structure (sometimes there are more than one building in a single picture; all of them are traced). The different damage levels used in the annotation task include:
\begin{itemize}
	\item \textcolor{black}{\it Severe (Fully Destroyed)}: A building should be considered fully destroyed if you see one or more of the following:
    \begin{itemize}
    	\item More than 50\% (half) of the building is damaged
    	\item 2 or more walls are destroyed
    	\item Roof is missing or completely destroyed
    \end{itemize}
    
    \item \textcolor{black}{\it Medium (Partially Damaged)}: A building should be considered partially damaged if you see one or more of the following:
    \begin{itemize}
    	\item About 30\% of the building is damaged
    	\item Damage to 1 or 2 walls, or 1 wall fully destroyed
    	\item Roof still there but perhaps damaged
    \end{itemize}
    
    \item \textcolor{black}{\it Mild (Little-to-no damage)}: A building should be considered little-to-no damage if 0\% to 10\% of the building is damaged.
    
\end{itemize}

\begin{figure}
   \centering
   \includegraphics[width=.9\columnwidth]{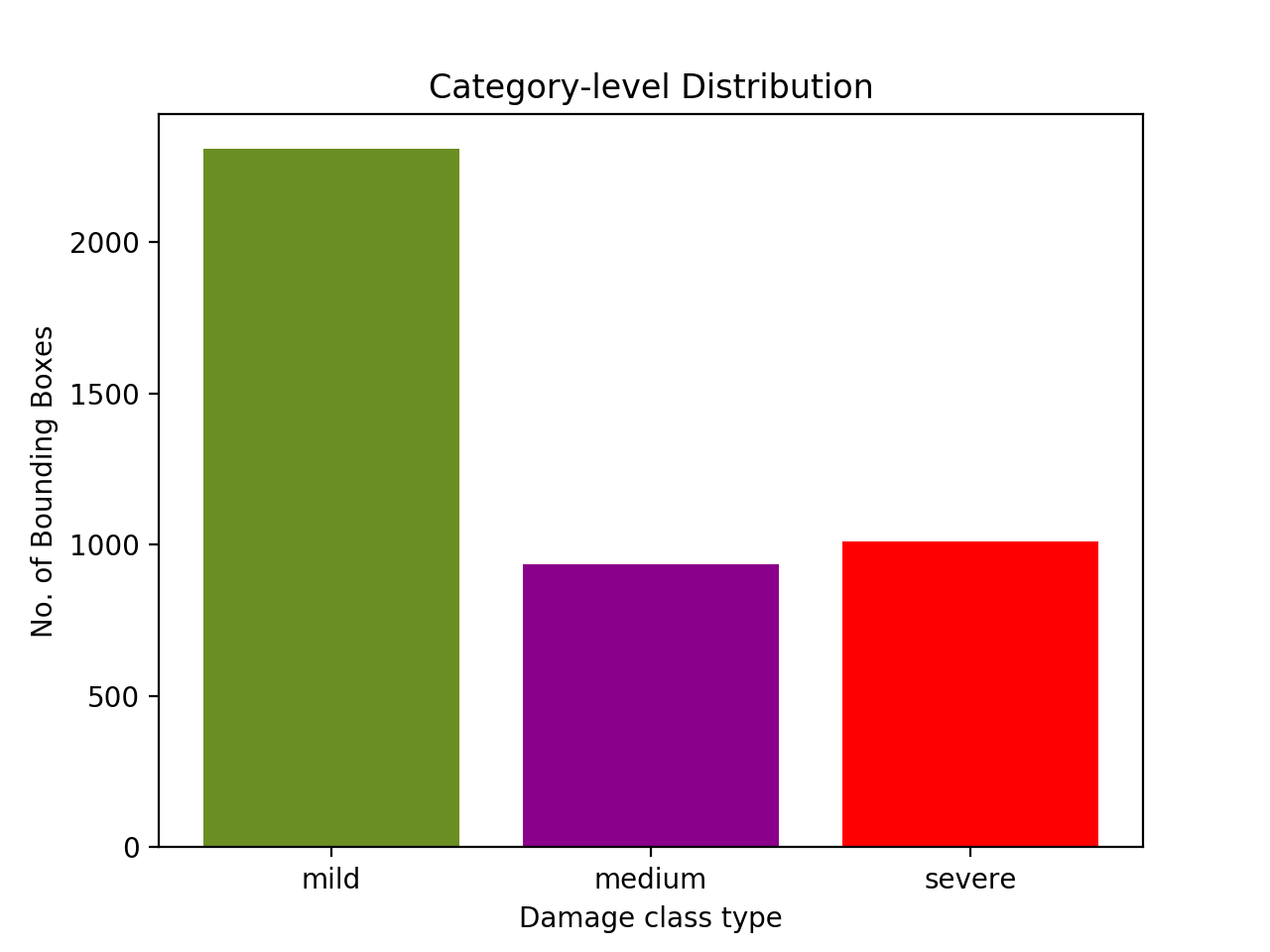}
   \caption{The bar chart shows the damage-class distribution among 4253 bounding boxes in a total of 1085 image dataset.}
   \label{fig:data_stat_dist}
\end{figure}

Thereafter, majority voting is done (from multiple annotations) to obtain the final label. Note that different buildings may be tagged with different damage levels in the same image. This annotation process resulted in a total of 2,979 segments corresponding to damaged structures (i.e., houses) in 1,085 images. Figure~\ref{fig:data_stat_dist} shows the class distribution, i.e. mild (54\%), medium (22\%) and severe (24\%). This appears to be situation of high class imbalance, almost half of the dataset is covered by mild category and the rest is divided between the other two difficult categories (almost) equally.
We perform 5-fold cross validation in all of the experiments and report performance in terms of mean and standard error.

\begin{figure*}[!t]
\centering
   \subfloat[]{\includegraphics[width=0.9\columnwidth]{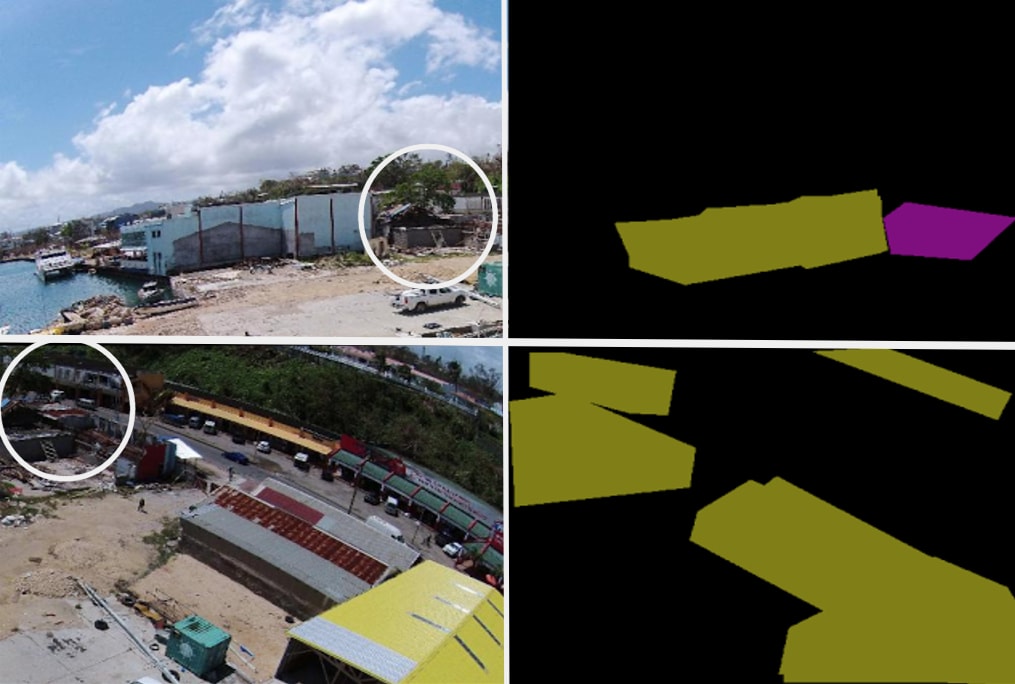}}
   \hfil
   \subfloat[]{\includegraphics[width=0.9\columnwidth]{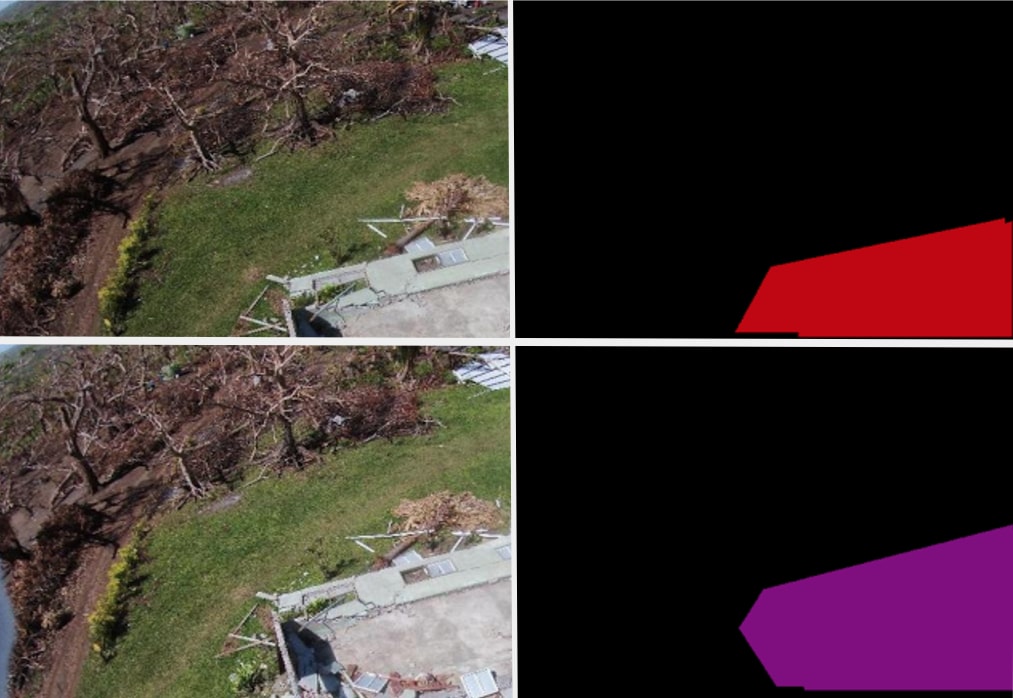}}
\caption{Few examples of disagreement in annotation: (a) Here, two images contain same house (circled) but with contrasting labels. (b) Another case where annotators agree on damage but on varied severity scale.}
\label{fig:labdis}
\end{figure*}

\subsubsection{Label Disambiguation} As mentioned earlier, the labels were annotated by crowd and there is a huge subjective incongruence which makes labels inconsistent thereby affecting learning. Also, the ground-truth is quite noisy since crowd-labeled polygons do not have crisp region boundaries. Figure~\ref{fig:labdis} highlights few cases where difference in annotators point of view hold true. We can clearly see coarse annotations for the second image from the top; and rest others show lack in labeling consensus. Due to this reason, the overall performance of the system is hampered to some extent. We now report on the  result of the first experiment.

\subsection{Evaluation of FV-CNN as a Baseline}
\label{ssec:fvcnn_baseline}
In this section we evaluate the performance of FV-CNN~\cite{cimpoi2015deep} for carrying out fine-grained damage assessment assuming that the ground-truth region is known and leveraging texture features using Fisher encoding. FV-CNN uses a pre-trained VGG-M model. The hidden layer used is a 512-dimensional local feature vector which gets further pooled into a Fisher Vector representation with a Gaussian Mixture Model (GMM) of size 64. The total resulting dimensionality is around 65K ($64 + 2\times64\times512$). We assume each component of the GMM has a diagonal covariance matrix. Finally, the region descriptors are classified using one-vs-all Support Vector Machines with a regularization hyperparameter $C=1$.

Despite the severe imbalance of damage type segments, FV encoding based classification gave good results across classes. The confusion matrix in Table~\ref{table:conf_3} show that regions with mild damage are classified with high accuracy. Medium and severe damage categories have lower accuracy. It is important to note that there was substantial disagreement between the annotators about the level of damage suffered by built structures and the accuracy was exacerbated due to the class-imbalance problem. We use FV-CNN for further analysis mainly for two reasons: (i) our problem is closely related to texture analysis, and (ii) FV encoding is quite powerful in discriminating objects based on texture, and has achieved state-of-the-art performance in several texture benchmarks~\cite{cimpoi2015deep}. FV-CNN experiments were run using the \textit{MatConvNet} toolbox on a Tesla K20Xm 6GB GPU card.

\begin{table}[!t]
\caption{\label{table:conf_3} Confusion Matrix for FV-CNN Damage Classification} 
\centering 
\begin{tabular}{l r r r r} 
\toprule 
GT vs. Pred & Mild & Medium & Severe \\ 
\midrule 
Mild & 84$\pm 0.88$ & 13$\pm 0.60$ & 3$\pm 1.17$ \\ 
Medium & 17$\pm 1.33$ & 69$\pm 1.18$ & 12$\pm 0.51$ \\
Severe & 8$\pm 1.50$ & 23$\pm 1.31$ & 67$\pm 0.73$ \\ 
\bottomrule 
\end{tabular}

\end{table}


\subsection{Pixel-level Segmentation by DeepLab as a Baseline}
\label{ssec:deeplab_baseline}
While the performance of FV-CNN for distinguishing between different damage levels based on texture is high, our aim is not only to  recognize the type of damage but also to identify regions in the image where the damage occurred. Therefore, FV-CNN cannot be a solution by itself for our problem because it assumes the availability of the segments (bounding boxes) at test time. For this reason, we used DeepLab~\cite{deeplab:iclr15}, a deep Convolutional Neural Network with a Conditional Random Field (CRF) layer on top, to segment the images so that the regions of potential damage can be recognized. The DeepLab system is one of the top-performing models on the PASCAL VOC-2012 semantic image segmentation task, reaching 71.6\% mean average precision (mAP) and has thus become a natural choice for semantic classification. We used the ``DeepLab-LargeFOV'' model to perform per-pixel classification for the given images generating segments belonging to one of the damage categories (or background). CRFs were used as post-processing step to discern local shapes and overcome the noisy labeling of the annotators. Additionally, to overcome the significant class-imbalance, we modified the cross-entropy loss function using class-weighting~\cite{badrinarayanan2015segnet}.

Table~\ref{table:dcnn_ss} shows the cross-validation results for our data with and without class weighting. Though, the original paper suggests \textit{mAP} is a better evaluation metric, however given the quality of annotations for our images (they are neither weak nor strong but somewhat of intermediate quality), we have used mean pixel-wise accuracy as a baseline for evaluating the results in \Naz. It is important to note that class-weighting enhances the localization and discriminative capability of the DeepLab system and improves the overall mean accuracy. DeepLab semantic segmentation experiments were run using the \textit{Caffe} framework on an NVIDIA K3100M 4GB GPU memory card with a batch size of 4.

\begin{figure}[!t]
    \centering
    \includegraphics[scale=0.36]{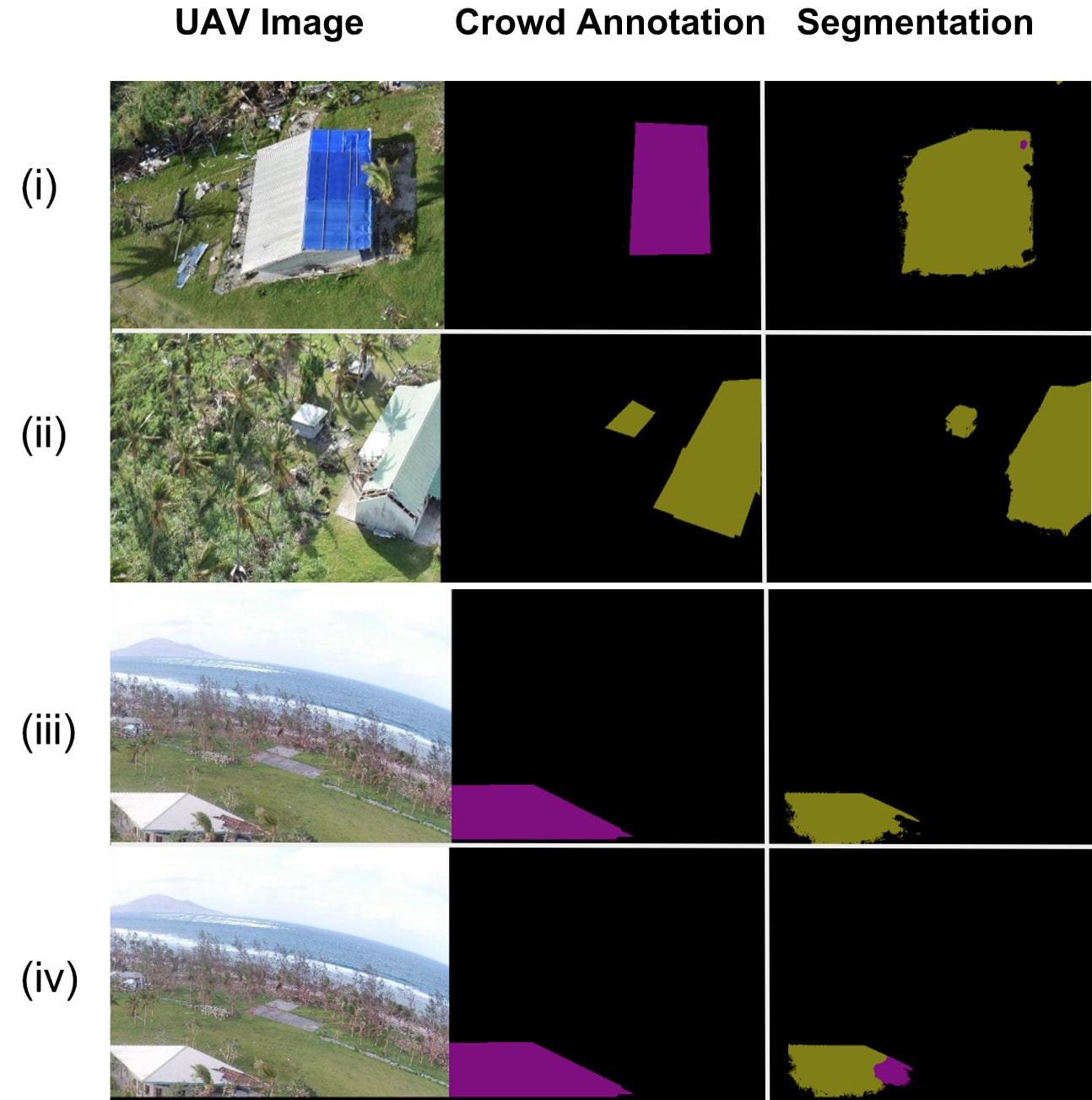}\\
    \caption{Semantic segmentation results by DeepLab~\cite{deeplab:iclr15}: (i)-(ii) shows that DeepLab is good at identifying damage regions but weak at texture discrimination; (iii)-(iv) reflects class-weighting enhances the localization and discriminative capability of the DeepLab system.}
    \label{fig:deeplab}
\end{figure}

In Figure~\ref{fig:deeplab} we have provided a few example images which highlight that DeepLab is good in identifying damage regions but poor at texture discrimination. Additionally, the poor performance is partly due to the fact that we have noisy ground-truth annotations.

\subsection{Proposed Pipeline}
\label{ssec:proposed_res}
\Naz\ combines pixel-level segmentation and fisher encoding. The accuracy of the combined system is shown in Table~\ref{table:ssfv}. The average accuracy of the DeepLab ($X$) system is shown in the first column. It is clear that class weighting improves the overall accuracy. In the second column the standalone accuracy of FV-CNN ($Y$) is shown. Thus if we combine the two pipelines, the hypothetical best accuracy is a product of $X$ and $Y$ which is shown in column three. However \Naz\ does slightly better then a simple product of the two. This suggests that in \Naz\ the FV-CNN component is able to fix some of the errors incurred by DeepLab. The overall performance of \Naz\ is limited by segments obtained by DeepLab; but DeepLab is affected due to poor annotations and multiple objects.

\begin{figure*}
\centering
  \subfloat[]{\includegraphics[width=0.72\textwidth]{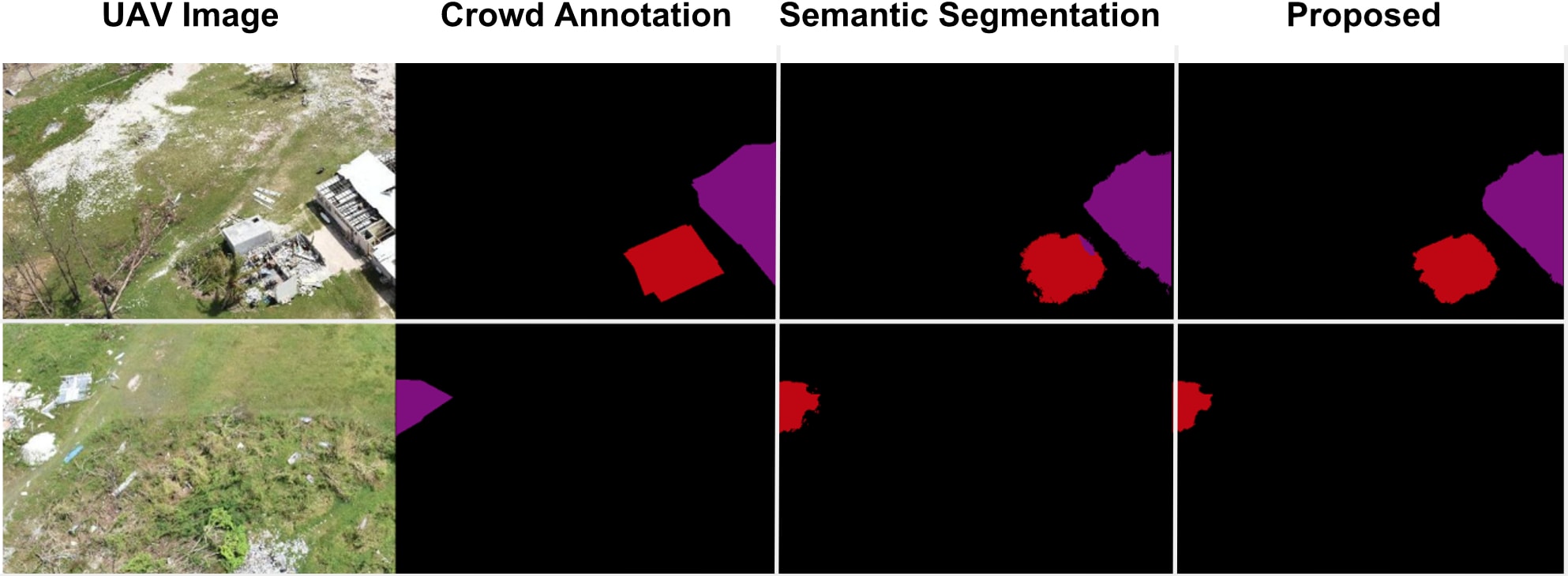}}\hfill
  \subfloat[]{\includegraphics[width=0.72\textwidth]{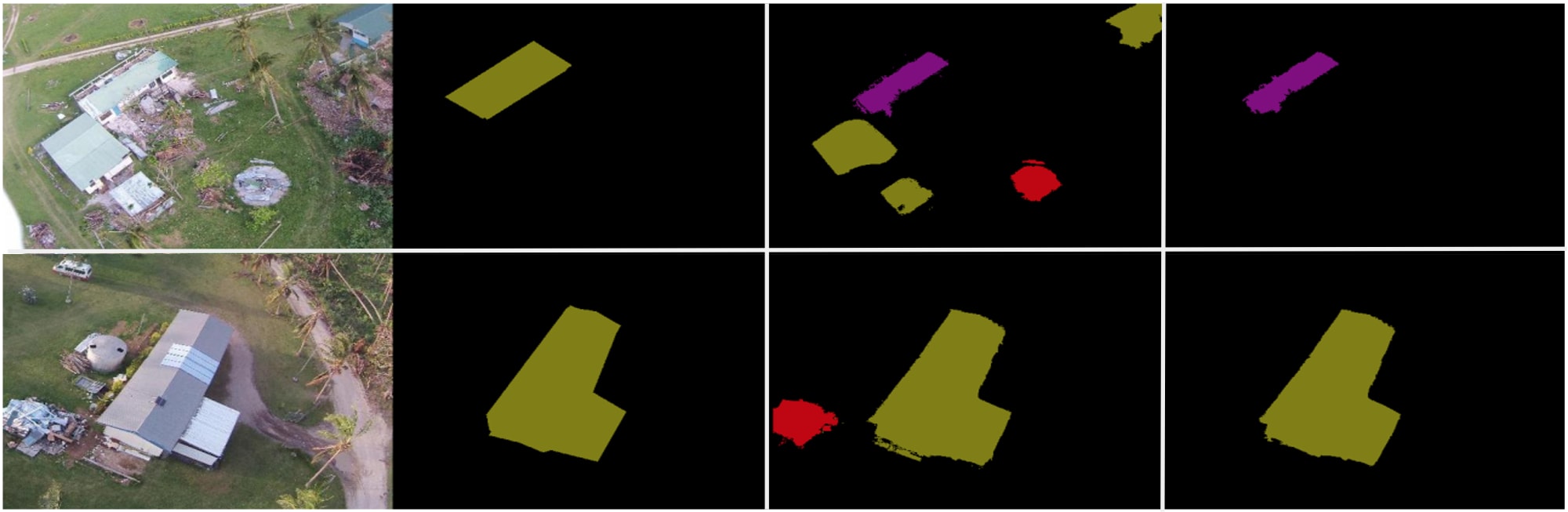}}\hfill
  \subfloat[]{\includegraphics[width=0.72\textwidth]{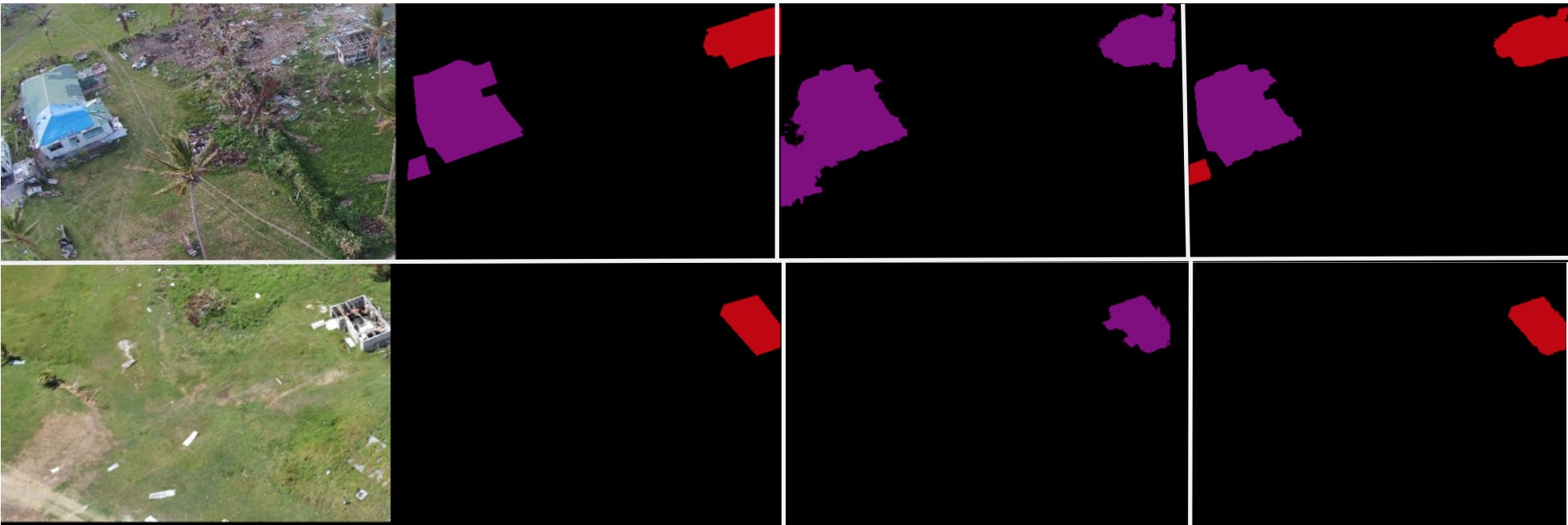}}\hfill
  \subfloat[]{\includegraphics[width=0.72\textwidth]{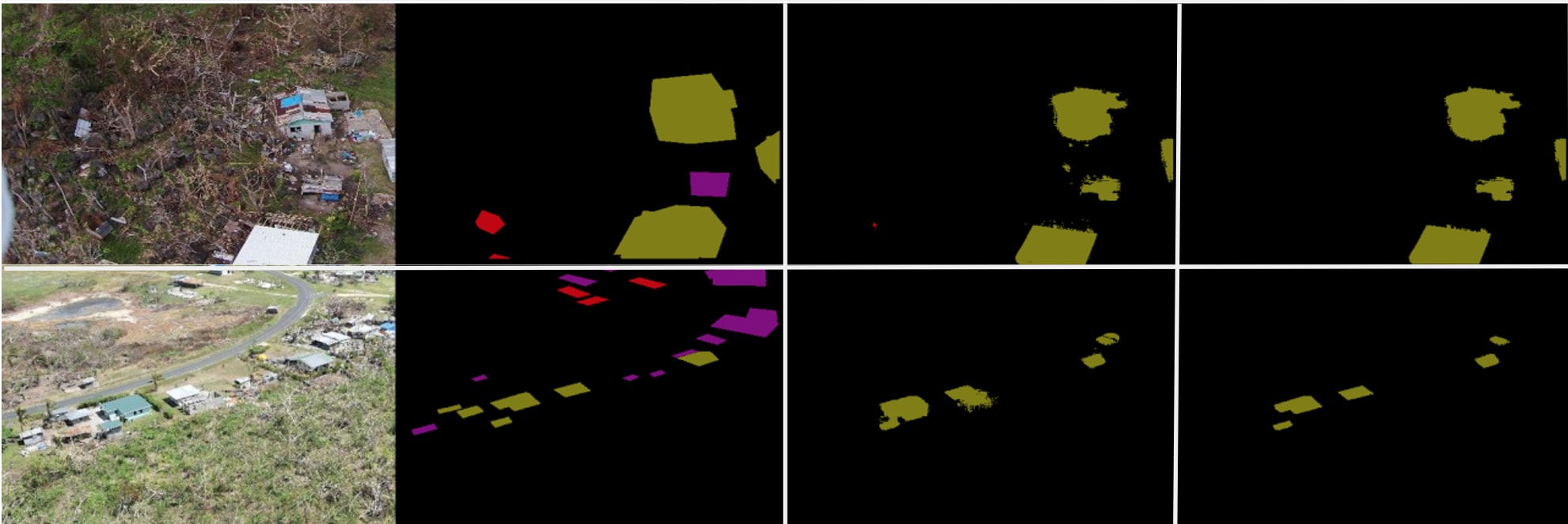}}\hfill
\caption{Examples to evaluate semantic segmentation~\cite{deeplab:iclr15} and our proposed pipeline (\Naz): (a) Two models are equally good. Also, the bottom image highlights that both algorithms are powerful in capturing the obvious severe damage in the UAV image which was originally labeled as medium, thereby reflecting label inconsistency. (b) Both examples show extra segments identified by semantic segmentation which are not annotated by crowd. This affects the evaluation due to improper ground truth in these cases. (c) \Naz\ performs well in differentiating between severe and medium. Learns texture in an efficient manner. (d) Hard cases in identifying more than approximately 10-15 objects in a single image.}
\label{fig:final_seg}
\end{figure*}

Figure~\ref{fig:final_seg} shows the detection and classification results on a few of the images with results from semantic segmentation and \Naz.
The examples shown in the figure highlight some of the following cases:
\begin{itemize}
      \item[(a)]{both models perform equally well,}
      \item[(b)]{labeling is poor but models are robust,}
      \item[(c)]{DeepLab is capable of identifying damage regions while Fisher Vector encoding helps in texture recognition.}
      \item[(d)]{the hard cases in identifying more than approximately 10-15 objects in a single image.}
\end{itemize}
   
The example images demonstrate that \Naz\ tends to combine the shape and texture features in a successful manner, and provides a reasonably robust results. In addition to reporting the average accuracy, we also compute average precision-recall values for semantic segmentation (DeepLab) and \Naz\ (DeepLab + FV-CNN) in Tables~\ref{tab:pr_deeplab} and~\ref{tab:pr_fvcnn}, respectively. We also compute the values with and without class weighting. The numbers show that precision drops with class weighting but recall improves for semantic segmentation. On the other hand, for \Naz\ precision values almost remains unchanged but again recall improves. In both tables, the achieved performance is lowest for the \textit{medium} damage category.

\begin{table}[!t]
\centering

\caption{\label{table:dcnn_ss} Semantic Segmentation Results by DeepLab~\cite{deeplab:iclr15}}\vspace{-2mm}
\begin{tabular}{lc} 
 \toprule
 Damage Level & Mean Accuracy(\%) \\ 
 \midrule
 3-class & 59.04$\pm 1.12$ \\ 
 3-class* & 63.07$\pm 1.71$ \\
 \bottomrule
 \multicolumn{2}{@{}l}{{\scriptsize * indicates training with class weighting}} \\
\end{tabular}

\bigskip

\caption{\label{table:ssfv} Comparison of all the models. \Naz\ performs well in differentiating texture, and improves on classifying segments identified by DeepLab}\vspace{-2mm}
\resizebox{\columnwidth}{!}{%
 \begin{tabular}{lcccc} 
 \toprule
 Damage & Segmentation & FV-CNN & Hypothesis & \Naz \\
 Levels	& (X) & (Y) & (X*Y) & \\
 \midrule
 3-classes & 59.04$\pm 1.12$ & 83.6$\pm 0.86$	 & 49.35$\pm 0.96$ & 50.25$\pm 1.51$  \\ 
 3-classes* & 63.07$\pm 1.71$ & 83.6$\pm 0.86$ & 52.71$\pm 1.47$ & 53.41$\pm 2.04$ \\ 
 \bottomrule
 \multicolumn{5}{@{}l}{{\scriptsize * indicates training with class weighting}} \\
\end{tabular}%
}

\bigskip

\caption{\label{tab:pr_deeplab} Precision-Recall for DeepLab per damage category}\vspace{-2mm}
\begin{tabular}{lcccc}
\toprule
Damage & \multicolumn{2}{c}{Precision} & \multicolumn{2}{c}{Recall} \\
Levels & w/o cls wt & w/ cls wt & w/o cls wt & w/ cls wt \\
\midrule
Mild   & 78.83$\pm 2.97$ & 66.87$\pm 1.49$ & 62.42$\pm 3.33$ & 73.90$\pm 4.20$  \\
Medium & 59.86$\pm 2.63$ & 50.67$\pm 3.19$ & 43.04$\pm 5.06$ & 43.70$\pm 5.72$  \\
Severe & 78.88$\pm 6.21$ & 67.76$\pm 7.05$ & 39.83$\pm 8.21$ & 48.85$\pm 4.76$  \\
\bottomrule
\end{tabular}

\bigskip

\caption{\label{tab:pr_fvcnn} Precision-Recall for \Naz\ per damage category}\vspace{-2mm}
\resizebox{\columnwidth}{!}{%
\begin{tabular}{lcccc}
\toprule
Damage & \multicolumn{2}{c}{Precision} & \multicolumn{2}{c}{Recall} \\
Levels & w/o cls wt & w/ cls wt & w/o cls wt & w/ cls wt \\
\midrule
Mild   & 86.02$\pm 5.06$ & 85.59$\pm 5.50$ & 64.34$\pm 2.64$ & 74.66$\pm 3.42$  \\
Medium & 56.38$\pm 6.98$ & 56.96$\pm 7.29$ & 30.44$\pm 11.46$ & 36.08$\pm 10.21$  \\
Severe & 79.24$\pm 11.21$ & 79.64$\pm 6.89$ & 38.85$\pm 8.68$ & 44.73$\pm 6.96$  \\
\bottomrule
\end{tabular}
}

\end{table}

\begin{figure*}[!t]
\centering
  \subfloat[]{\includegraphics[width=0.72\textwidth]{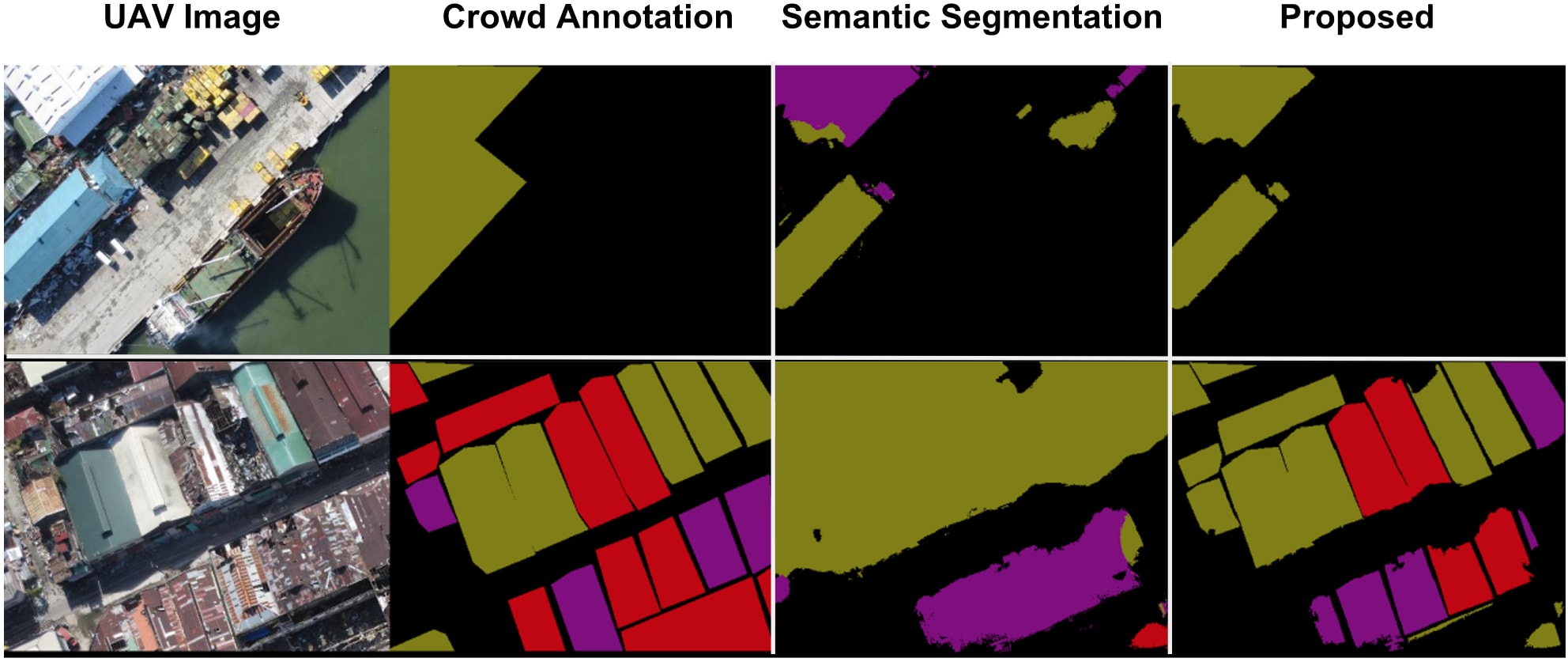}}
  \hfil
  \subfloat[]{\includegraphics[width=0.72\textwidth]{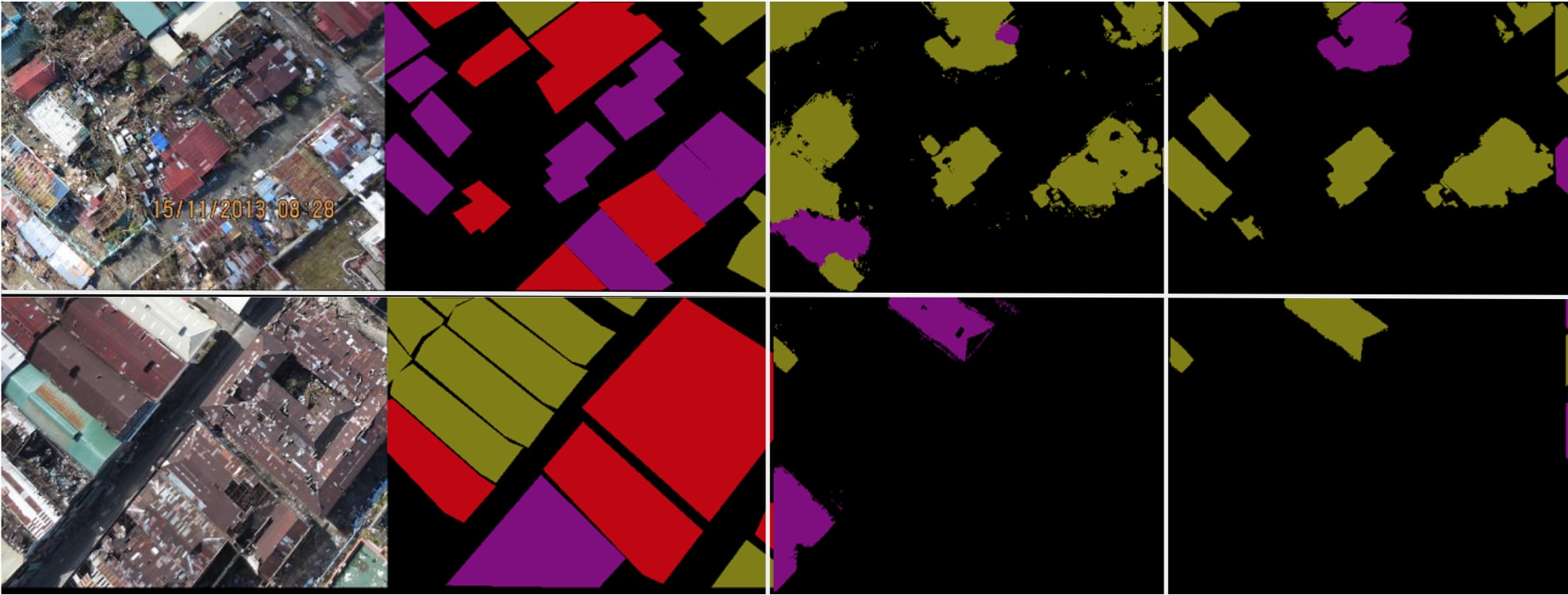}}
  \caption{\label{fig:philly} Transfer learning on Philippines data: (a) Both images show that our proposed pipeline (\Naz) can discriminate texture to some degree. (b) Exhibit the difficult case (top image) as well as poor prediction (bottom). The type of damage in this case is quite different and more difficult to categorize; often the roofs are also of same color and material type.}
\end{figure*}

\subsection{Transfer Learning}
\label{ssec:transfer_res}
To test the effectiveness of \Naz\ and the need to create the high-performance learner for a target domain from a similar source domain, we performed transfer learning. We evaluated our model on the 60 images of disaster-struck (typhoon) areas in Philippines. The terrain in Philippines is quite different from that of our original data set; it is more like a city landscape. The results in Table~\ref{table:transfer_learning} show that \Naz\ performs relatively better and improves on mild and severe classes than the baseline semantic segmentation (DeepLab). This is further exhibited in Figure~\ref{fig:philly} which shows the texture discriminatory potential to some degree. It also draws attention to the cases where the models perform poorly due to the fact that the damage is observed mostly on the roofs with same color and differentiating medium from severe proves to be very challenging. Furthermore, the low number for severe category is mainly due to imbalanced data as most of the buildings come under mild category.

\begin{table}[!t]
\caption{\label{table:transfer_learning} Evaluation of transfer learning on Philippines data}
\centering
\begin{tabular}{lcccc} 
 \toprule
Model & Overall &  \multicolumn{3}{c}{Per-class Accuracy (\%)} \\
Architecture & Accuracy (\%) & Mild & Medium & Severe \\ 
\midrule
 Segmentation & 36.26 & 62.80 & 35.70 & 10.30 \\
 \Naz\ & \textbf{40.90} & 86.77 & 17.01 & 16.50 \\ 
 \bottomrule
\end{tabular}
\end{table}

\newcommand{\Ni}{({\em i})~}
\newcommand{\Nii}{({\em ii})~}
\newcommand{\Niii}{({\em iii})~}
\newcommand{\Niv}{({\em iv})~}
\section{Related Work}
\label{sec:rel_work}

Deep Learning techniques now underpin many computer vision tasks. In particular the best algorithms for
image classification are now likely to be based on  Convolutional Neural Networks (CNNs)~\cite{yann:org_ppr,
NIPS2012_4824, sermanet-iclr-14, SimonyanZ14a, girshick2014rcnn, SzegedyC:CVPR15, He_2016_CVPR}. CNNs (especially
with the pooling layer) are designed to be translation invariant. However one of the key strengths of
CNNs, that they are designed to be invariant to spatial transformations, is precisely the weakness when it comes
to object localization as required for damage assessment from UAVs. To overcome the weakness of CNNs,
Chen et.al.~\cite{deeplab:iclr15}, have introduced the ``DeepLab'' system which combines CNNs with Conditional
Random Fields (CRFs). CRFs are particularly useful for capturing local interaction between
neighboring pixels. In particular DeepLab performs pixel-level classification, a task sometimes known
as semantic image segmentation  (SS) in the computer vision community. 
Early attempts in semantic image segmentation used a set of bounding boxes and masked regions as input to the CNN architecture to incorporate shape information into the classification process to perform object localization and semantic segmentation~\cite{girshick2014rcnn, Hariharan:ECCV14, Mostajabi:CVPR15, Girshick:2015vr,Ren:2015ug}. Taking a slightly different approach, some studies employed segmentation algorithms independently on top of deep CNNs that were trained for dense image labeling~\cite{Hariharan_2015_CVPR, Farabet_PAMI13}. More direct approaches, on the other hand, aim to predict a class label for each pixel by applying deep CNNs to the whole image in a fully convolutional fashion~\cite{Shelhamer:TPAMI16,Eigen:ICCV15}. Similarly, \cite{badrinarayanan2015segnet} and \cite{Noh_2015_ICCV} train an end-to-end deep encoder-decoder architecture for multiclass pixelwise segmentation.

One of the key strengths of systems based on deep learning is that they  automatically infer a representation of the data suitable
for the defined task. For example, the lower layers of deep learning correspond to a representation 
suitable for low-level vision tasks while the higher layers are more domain specific~\cite{goodfellow} and
obviates the need for pre-defined feature engineering like SIFT~\cite{Lowe:sift}. Fisher Vectors (FV) are
an important representation used in computer vision for object discrimination~\cite{sanchez2013image}. 
FVs generalize the Bag of Visual Word (BoV) model which are now often built on top of CNN hidden layers. In
particular  FV-CNN~\cite{cimpoi2015deep}, is a recent attempt to combine the use of CNN and FVs for texture recognition and segmentation.

Aerial image analysis for detecting objects, classifying regions and analyzing human behavior is an active research area. A recent overview is presented in Mather and Koch \cite{mather2011computer} which also mentions the use of damage assessment datasets (e.g.,~\cite{DamAssessData}) as a benchmark. Works which use texture for aerial imagery include ~\cite{HungC:RS14}. Examples of other recent work include Blanchart et al.~\cite{Blanchart:ICIP11}, where they utilize SVM-based active learning to analyze aerial images in a coarse-to-fine setting. Bruzzone and Prieto~\cite{Bruzzone:TIP02} is an example of a change detection-based analysis technique. Zhang et al.~\cite{ZhangH:ICIP15} develop coding schemes for classifying aerial images by land use. Similarly, Hung et al.~\cite{HungC:RS14} tackle with weed classification based on deep auto-encoders while Quanlong et al.~\cite{QuanlongF:RS15} analyze the urban vegetation mapping using random forests with texture-based features. Oreifej et al.~\cite{OreifejO:CVPR10} recognize people from aerial images. Gleason et al.~\cite{GleasonJ:ICRA11} and  Moranduzzo and Melgani ~\cite{MoranduzzoT:TGRS14} detect cars in aerial images using kernel methods and support vector machines.

There are also studies that aim to produce a complete semantic segmentation of the aerial image into object classes such as building, road, tree, water~\cite{DollarP:CVPR06, KlucknerS:ICCVW09, KlucknerS:ACCV10}. Some of the recent attempts apply deep CNNs to perform binary classification of the aerial image for a single object class~\cite{DollarP:CVPR06, MnihV:ECCV10, MnihV:ICML12}. These recent attempts to apply deep learning techniques to high-resolution aerial imagery have resulted in highly accurate object detectors and image classifiers, suggesting that automated aerial imagery analysis systems may be within reach.

However, most of the aforementioned aerial image analysis methods assume that the images are captured at a nadir angle via satellites with known ground resolution, and hence, fixed viewpoint and scale for the objects in the scene. However, UAVs usually fly at variable altitudes and angles, and therefore, capture oblique images with varying object sizes and appearances. Therefore, in contrast to the traditional aerial image analysis and computer vision paradigms, a new set of computer vision and machine learning approaches must be developed for UAV imagery to account for such differences in the acquired image characteristics.

Finally, there are a few other recent damage assessment studies on images collected from social media platforms, not necessarily UAV images though. For example, \cite{petersinvestigating,daly2016mining} tackle with fire detection whereas \cite{LagerstromR:Frontiers16} addresses flood detection from social media images. Furthermore, \cite{alam17demo} proposes an automatic image processing pipeline for social media imagery data, and \cite{nguyen2017automatic, nguyen17damage} further explore infrastructural damage assessment problem, mainly at the image classification level.

\section{Summary and Future Work}
\label{sec:conclusion}
In this paper we have proposed an integrated deep learning pipeline (\Naz) for identifying built structures (e.g., houses) in UAV images followed by a fine-grained damage classification. The images were collected in the aftermath of a natural disaster with the aim of assessing the level of damage. 

\Naz\ has two distinct components. The first component carries out a pixel-level classification of images (a task often known as semantic segmentation) with the aim of identifying damaged structures in an image. The aim of the second component is to carry out a fine-grained classification of the structures identified to assess the severity of damage. We use a Fisher Vector representation of image segments to assess the severity of damage. \Naz\ is particularly robust against noisy labels and appears to be height invariant--a necessary property for UAV images.

To the best of our knowledge this is the first known deep learning pipeline for object detection and classification of UAV images collected from disaster struck regions. 
At this point, our work handles a more complex problem, such as, image segmentation for noisy aerial images. We plan to further investigate end-to-end deep learning techniques to better handle the noise and ambiguity in ground truth annotations.







\bibliographystyle{IEEEtran}
\balance
\bibliography{egbib}
%


\end{document}